\documentclass[sigconf]{acmart}
\usepackage{subfigure}
\usepackage{amsmath}
\usepackage{threeparttable}
\usepackage{soul}    
\soulregister\cite7     
\soulregister\ref7 
\def\BibTeX{{\rm B\kern-.05em{\sc i\kern-.025em b}\kern-.08emT\kern-.1667em\lower.7ex\hbox{E}\kern-.125emX}}

\makeatletter
\def\MT@is@composite#1#2\relax{%
	\ifx\\#2\\\else
	\expandafter\def\expandafter\MT@char\expandafter{\csname\expandafter
		\string\csname\MT@encoding\endcsname
		\MT@detokenize@n{#1}-\MT@detokenize@n{#2}\endcsname}%
	\ifx\UnicodeEncodingName\@undefined\else
	\expandafter\expandafter\expandafter\MT@is@uni@comp\MT@char\iffontchar\else\fi\relax
	\fi
	\expandafter\expandafter\expandafter\MT@is@letter\MT@char\relax\relax
	\ifnum\MT@char@ < \z@
	\ifMT@xunicode
	\edef\MT@char{\MT@exp@two@c\MT@strip@prefix\meaning\MT@char>\relax}%
	\expandafter\MT@exp@two@c\expandafter\MT@is@charx\expandafter
	\MT@char\MT@charxstring\relax\relax\relax\relax\relax
	\fi
	\fi
	\fi
}
\def\MT@is@uni@comp#1\iffontchar#2\else#3\fi\relax{%
	\ifx\\#2\\\else\edef\MT@char{\iffontchar#2\fi}\fi
}
\makeatother

\setcopyright{acmcopyright}
\begin{document}
	
	%
	\title{Walking with MIND: \underline{M}ental \underline{I}magery e\underline{N}hance\underline{D} Embodied QA}
	
	%
	
		\author{Juncheng Li}
		\affiliation{%
			\institution{Zhejiang University}
		}
		\email{jethro9783@gmail.com}

		\author{Siliang Tang}
		\authornote{Siliang Tang is the correspondence author.}
		\affiliation{%
			\institution{Zhejiang University}
		}
		\email{siliang@zju.edu.cn}

		\author{Fei Wu}
		\affiliation{%
			\institution{Zhejiang University}
		}
		\email{wufei@zju.edu.cn}
		
		\author{Yueting Zhuang}
		\affiliation{%
			\institution{Zhejiang University}
		}
		\email{yzhuang@zju.edu.cn}
	%
	\renewcommand{\shortauthors}{Juncheng Li and Siliang Tang, et al.}
	%

	\begin{abstract}
		The EmbodiedQA is a task of training an embodied agent by intelligently navigating in a simulated environment and gathering visual information to answer questions. Existing approaches fail to explicitly model the mental imagery function of the agent, while the mental imagery is crucial to embodied cognition, and has a close relation to many high-level meta-skills such as generalization and interpretation. In this paper, we propose a novel \underline{M}ental \underline{I}magery e\underline{N}hance\underline{D} (\textbf{MIND}) module for the embodied agent, as well as a relevant deep reinforcement framework for training. The MIND module can not only model the dynamics of the environment (\textsl{e.g. `what might happen if the agent passes through a door'}) but also help the agent to create a better understanding of the environment (\textsl{e.g. ‘The refrigerator is usually in the kitchen’}). Such knowledge makes the agent a faster and better learner in locating a feasible policy with only a few trails. Furthermore, the MIND module can generate mental images that are treated as short-term subgoals by our proposed deep reinforcement framework. These mental images facilitate policy learning since short-term subgoals are easy to achieve and reusable.  This yields better planning efficiency than other algorithms that learn a policy directly from primitive actions. Finally, the mental images visualize the agent's intentions in a way that human can understand, and this endows our agent's actions with more interpretability. The experimental results and further analysis prove that the agent with the MIND module is superior to its counterparts not only in EQA performance but in many other aspects such as route planning, behavioral interpretation, and the ability to generalize from a few examples. 
		
	\end{abstract}
\copyrightyear{2019} 
\acmYear{2019} 
\acmConference[MM '19]{Proceedings of the 27th ACM International Conference on Multimedia}{October 21--25, 2019}{Nice, France}
\acmBooktitle{Proceedings of the 27th ACM International Conference on Multimedia (MM '19), October 21--25, 2019, Nice, France}
\acmPrice{15.00}
\acmDOI{10.1145/3343031.3351017}
\acmISBN{978-1-4503-6889-6/19/10}
	
	%
	%
	
	%

	
	%
	
	\maketitle

	\vspace{-0.1cm}
	\section{INTRODUCTION}
\begin{figure}[!t]
		\centering
		\includegraphics[width=\linewidth]{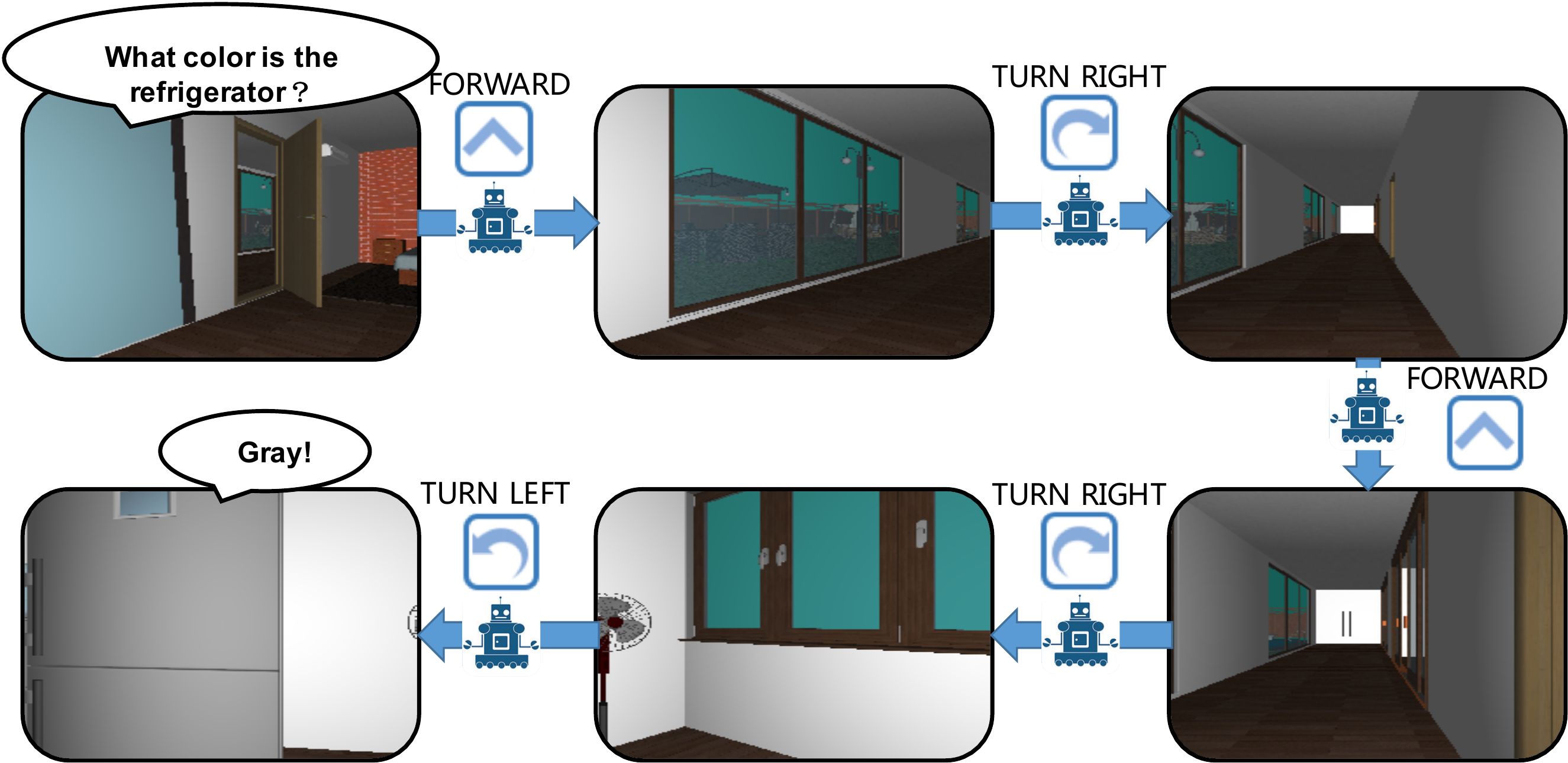}
		\caption{An example of Embodied QA task.} 
		\label{f0}
		\vspace{-0.4cm}
	\end{figure}

	The EmbodiedQA~(EQA), proposed by~\cite{Das_2018_CVPR}, is a task of training an embodied agent which is required to intelligently navigate in a simulated environment and gather visual information to answer questions.  For example, as shown in Figure~\ref{f0}, the agent is initialized at a random location and asked a question (\textsl{e. g. `What color is the refrigerator?'}). In order to answer the question, the agent has to explore the environment and find a visually grounded answer to the question. Different from all previous vision-language tasks, such as Image Captioning and Visual QA, embodied means that the environment is part of the cognitive system~\cite{wilson2002six} that will influence the mind. This indicates the cognition and the actions are based on the understanding of the dynamics of the environment as well as common sense knowledge~(\textsl{e.g. ‘we usually enter the room from the door’})  that gradually collected from daily experiences.  Therefore, a mental model~\cite{johnson1995mental, jones2011mental} with imagery function~\cite{finke1989principles, zalta2003stanford, kosslyn1994image}  that models the environment is crucial to building an embodied agent. 
	
	Many deep reinforcement learning based methods~\cite{Das_2018_CVPR, das2018neural, gordon2018iqa} have been proposed to improve the EQA task. They are designed to achieve the ultimate objectives by learning to select primitive actions without long-term planning. None of them explicitly model the mental imagery function of the agent, while the mental imagery is important to long-term planning and has a close relation to many valuable high-level meta-skills such as generalization and interpretation.
	
	When humans are solving a task, they do not make actions solely based on current observations. There is a mental imagery model behind every decision. The model comes from our everyday experiences; it predicts the dynamics of the environment and forms some mental images (i.e. mental imagery) in our mind.  These mental images can be viewed as short-term subgoals that provide a path to a more concrete solution. Take building a house with Lego for example. Firstly, we may imagine the foundation of the house and build the foundation based on the imaginary `sketch'. Then we may imagine the main body and the roof of the house in turn, and finally, build the whole house according to the imagery in our minds. In this case, we are not building the house directly without planning, but dividing it into several sub-stages (i.e. subgoals) and imagining the scene of each individual stage. Mental imagery facilities the understanding of our physical world;  it can predict the future without having to experience that outcome directly. It is a meta-skill that help us to achieve different tasks across different domains more efficiently.
	
	Motivated by the role of the mental imagery in human's decision making, we propose a novel \underline{M}ental \underline{I}magery e\underline{N}hance\underline{D} (\textbf{MIND}) module for the embodied agent, as well as a relevant deep reinforcement framework for training. The MIND module is composed of a Mental Autoencoder and an Imagery Model. Together they explicitly model the dynamics of the environment~(\textsl{e.g. `what might happen if the agent passes through a door'}). An agent with MIND module will have several advantages: 
	\vspace{-0.1cm}
	\begin{itemize}
		\item \textbf{Faster Convergence:} The MIND module helps the agent to create a better understanding of the environment (\textsl{e.g. ‘The refrigerator is usually in the kitchen’}). Such knowledge makes the agent a faster and better learner in locating a feasible policy with only a few trails. 
		\item \textbf{Better Generalizability:} Since the MIND module explicitly models the dynamics of the environment, the learned module is transferable across different tasks, as long as the dynamics of the environment remains the same. This is especially useful for learning in unknown scenes with only a few training examples are available. 
		\item \textbf{Better Planning Efficiency:} The MIND module can generate mental images that are treated as short-term subgoals by our proposed deep reinforcement framework. In this new RL framework, we designed a special reward (i.e. \textbf{planned reward}) to encourage our agent learning to form more task-related and objective-related short-term subgoals~(\textsl{e.g. `leave the room', `go down the corridor', etc.}). These subgoals are easy to achieve and reusable.
		\item \textbf{Better Behavioral Interpretation:} The mental images generated in planning time visualize the agent's intentions in a way that human can understand, and this makes real-time behavioral interpretation or even correction feasible. 
	\end{itemize}

	
	
	
	\vspace{-0.1cm}
	\section{RELATED WORK} \label{s2}
	
	
	\subsection{Embodied Question Answering}
	Recently, several deep reinforcement learning based hierarchical architectures for EmbodiedQA have been proposed by Gordon \textsl{et al.}  \cite{gordon2018iqa} in the AI2-THOR environments \cite{kolve2017ai2}, and by Das \textsl{et al.} \cite{Das_2018_CVPR, das2018neural} in the House3D environments \cite{wu2018building}. These approaches decompose the control problem into multiple levels and consist of a factorized set of modules \cite{tessler2017deep, andreas2017modular, oh2017zero}. Gordon \textsl{et al.} \cite{gordon2018iqa} propose the Hierarchical Interactive Memory Network (HIMN), consisting of a high-level planner and some low-level controller, allowing the agent to operate at multiple levels of temporal abstraction. The high-level planner chooses the task to be performed 
	 and the speciﬁed low-level controller 
	 executes the task. Das \textsl{et al.} \cite{Das_2018_CVPR} divide EmbodiedQA agent into four modules-vision, language, navigation and answering, and the navigation module (PACMAN) decomposes navigation into a planner, which selects actions 
	  to perform, and a controller, which performs these actions. Das \textsl{et al.} \cite{das2018neural} later propose a hierarchical Neural Modular Controller (NMC), consisting of a master policy and several sub-policies. 
	  These approaches all ignore the crucial importance of the mental imagery model on embodied cognition, which results in poor generalizability and low planning efficiency. More specifically, they do not consider the dynamics of the environments, making it harder to generalize to new scenes. Besides, most approaches just execute primitive actions over long time horizons. They can not plan to complete a sequence of short-term subgoals and finally answer the questions. Although NMC \cite{das2018neural} contains a master policy to choose high-level subgoals,
	it requires additional training data annotated with a series of subgoals to train the master policy, and the types of subgoals are pre-defined, which is not useful in practice. In comparison, MIND module first models the environment dynamics, which enhances its generalizability, and predicts imaginary short-term subgoals, which guarantees the agent's planning efficiency and provides a path to visualize the agent’s intentions.
	
	\subsection{Mental Imagery}
	Mental imagery (varieties of which are sometimes colloquially referred to as`visualizing', `seeing in the mind's eye', `hearing in the head', etc.) is quasi-perceptual experience \cite{finke1989principles, zalta2003stanford, kosslyn1994image}; it resembles perceptual experience, but occurs in the absence of the appropriate external stimuli \cite{zalta2003stanford}. Numerous experiments carried out over the past twenty years have probed the nature of mental imagery and unlocked its powers \cite{finke1989principles, kosslyn1994image, ha2018world}. The predictive model in our minds which forms the mental imagery is called as mental model \cite{johnson1995mental, jones2011mental}. Ha \textsl{et al.} \cite{ha2018world} instantiate the mental model as `world model' and apply it to some games-Car Racing and VizDoom. However, it can not do long-term planning. It just predicts the latent representation of next frame after a primitive action, which has nothing to do with the task. While we desigh a RL framework to encourage our MIND module to generate task-related and interpretable prediction after several actions.
	

	\subsection{Vision-and-Language Navigation}
	Vision-and-language navigation(VLN) \cite{Anderson_2017} requires the agent to understand natural-language navigation instructions and achieve the ultimate goal in a simulated environment. Natural language command of robots in unstructured environments has been a research goal for several decades \cite{Win_1971}. Early approaches \cite{chen:aaai11,chaplot2017gatedattention,Macmahon06walkthe,mei2015listen} simplify the problem of visual perception to some degree. They restrict environments to require limited perception or enumerated all navigation goals or objects, and the navigation goal in these approaches is usually directly annotated in a prior global map. In recent work, Mei \textsl{et al.} \cite{mei2015listen} propose a neural sequence-to-sequence model to map the natural-language navigation instructions to actions. Anderson \textsl{et al.} \cite{Anderson_2017} formulate VLN as visually grounded sequence-to-sequence transcoding problem, and propose a sequence-to-sequence architecture with an attention mechanism, as well as a Room-to-Room dataset which is the first benchmark dataset in real buildings. Wang \textsl{et al.} \cite{wang2018reinforced} propose a Cross-Modal Matching Critic to reconstruct the language instructions from the trajectories executed by the navigator, which is aimed to encourage the global matching between them. Similar to EmbodiedQA, VLN  also needs to navigate in the environment to achieve some goals. The crucial difference between them is how the goals are speciﬁed. VLN explicitly provides a sequence of instructions and specifies the target. The VLN agent just needs to map the natural-language navigation instructions to actions. In contrast, EmbodiedQA does not provide instructions to the agent, and the ultimate goals are implicit in natural-language questions. The instructions in VLN can be seen as short-term subgoals, which require the agent to plan them by itself in EmbodiedQA.
	
	\section{Mental Imagery Enhanced Module} \label{s3}
	The MIND Module has two components: the Mental Autoencoder and the Imagery Model. The mental autoencoder receives raw RGB images through a single egocentric RGB camera  and learns a compressed spatial and temporal representation from the environment. As the agent explores the environment, it gets a series of images and has its mental representations. Then the imagery model can use the sequences of mental representations to form useful hypothesis of how the environment works and predicts the future without having to experience that outcome directly. 
	
	\subsection{Mental Autoencoder}
	The Mental Autoencoder is composed of a mental encoder and a mental decoder. 
	The mental encoder is aimed to extract spatial and temporal information from the environment and uses them to form mental representations. We use $\beta$-VAE \cite{higgins2017beta, beta_vae,vae} to discover  disentangled latent factors. This means each dimension of the inferred latent representation represents one single generative factor(e.g.,room direction,scale) and relatively invariant to other dimensions. Such disentangled representation has good interpretability and easy generalization to a variety of tasks. It enables the MIND module to control the imagery generation in a more interpretable way(e.g., increase the dimension that controls the room direction to generate an image that turns left from current view). 
	
	
	Formally, let $I_t$ denote $224\times224\times3$ raw RGB image, $m_t$ denote the mental representation encoded by the mental encoder, and $M_t$ denote the mental image produced by the mental decoder. 
	
\begin{figure}[h]
		\centering
		\includegraphics[width=\linewidth]{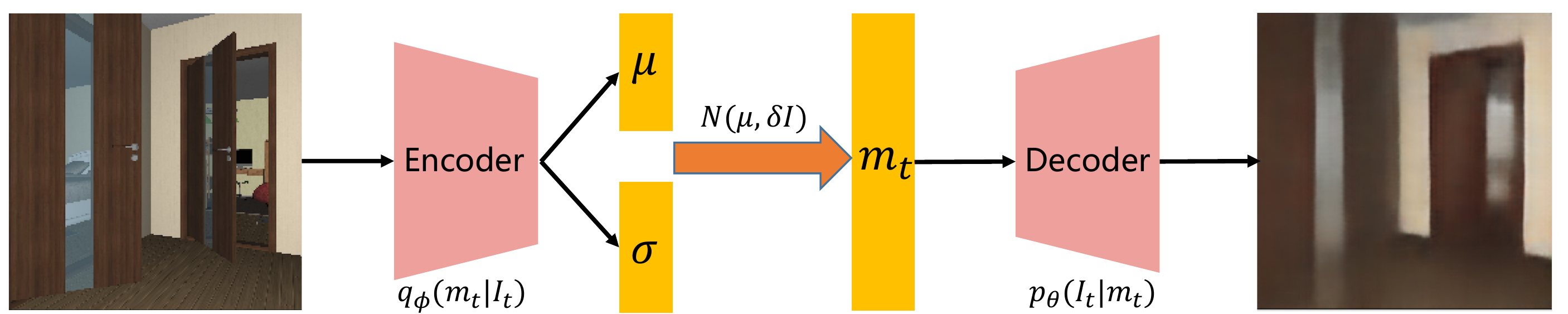}
		\caption{The encoder outputs low dimension vectors $\mu$ and $\delta$ which are the parameters of Gaussian distribution N(µ, σI). The decoder receives mental representation $m_t$ sampled from N(µ, σI) and uses it to reconstruct the original image. }
		\label{f1}
	\end{figure}

	As shown in Figure \ref{f1}, the encoder is a Convolutional Neural Networks (CNNs) \cite{krizhevsky2012imagenet} which takes $I_t$ as input and passes $I_t$ through 4 convolutional layers to encode it into low dimension vectors $\mu$ and $\delta$, with the same size $N_{m}$. The mental representation $m_t$ is then sampled from the Gaussian prior $N(\mu, \delta)$. The decoder is also instantiated as a neural network that learns to reconstruct the image given $m_t$. The loss function of $\beta$-VAE is defined as:
	\begin{gather}
	J_{vae} = -E_{m_t \sim q_{\phi}(m_t|I_t)}logp_{\theta}(I_t|m_t) \\
	\qquad  \quad  + \beta D_{KL}(q_{\phi}(m_t|I_t)||p_{\theta}(m_t))\notag
	\end{gather}
	where $\phi$ is the parameters of encoder and $\theta$ is the parameters of decoder.
	\vspace{-0.4cm}
	
	\subsection{Imagery Model}
	Through the mental encoder, we get a series of robust and disentangled mental representations in the process of agent exploring the environment. The imagery model is aimed to compress temporal information over time and predict the next mental representation $m_{t+1}$ given a specific action. Then we can decode $m_{t+1}$ by the mental decoder to get the mental image $M_{t+1}$ which reflects the imagery in mind. We use long short-term memory (LSTM) \cite{hochreiter1997long} for time series modelling and combine it with a Mixture Density Network (MDN) \cite{bishop1994mixture} as the output layer. Let $m_{t+1}^{'}$ denote the prediction of the next mental representation, distinguished from the real mental representation encoded by mental encoder at time t+1. Instead of a deterministic prediction of $m_{t+1}^{'}$, the MDN outputs the parameters of the mixture distribution $P(m_{t+1}^{'}|m_{t}, h_{t}, a_{t})$ and uses it to sample a prediction of the next mental representation $m_{t+1}^{'}$. Importantly, its mechanism for generating the next mental representation $m_{t+1}^{'}$ is similar to the mechanism of mental encoder (they both output the parameters of Gaussian distribution and sample mental representation from it).
	
	Formally, let $a_{t} \in \{forward, turn\ left, turn\ right, stop\}$ denote the action the agent will take, and $h_{t}$ denote LSTM's hidden state at time $t$. The imagery model predicts $m_{t+1}^{'}$ as follows:
	\begin{equation}
	m_{t+1}^{'}, h_{t+1} \leftarrow Imagery(m_{t}, h_{t}, a_{t})
	\end{equation}
	More specially, the MDN takes the LSTM's output $h_t$ as its input, outputs the parameters of a mixture of Gaussian distribution, and then samples $m_{t+1}^{'}$ from this distribution. Thus we can consider that the LSTM's hidden state contains the spatial and temporal information of the environment. When we combine the MIND Module with navigation model, we will use the LSTM's hidden state directly during planning. The details will be described in Section \ref{s4}.
	
\begin{figure}[h]
		\centering
		\includegraphics[width=\linewidth]{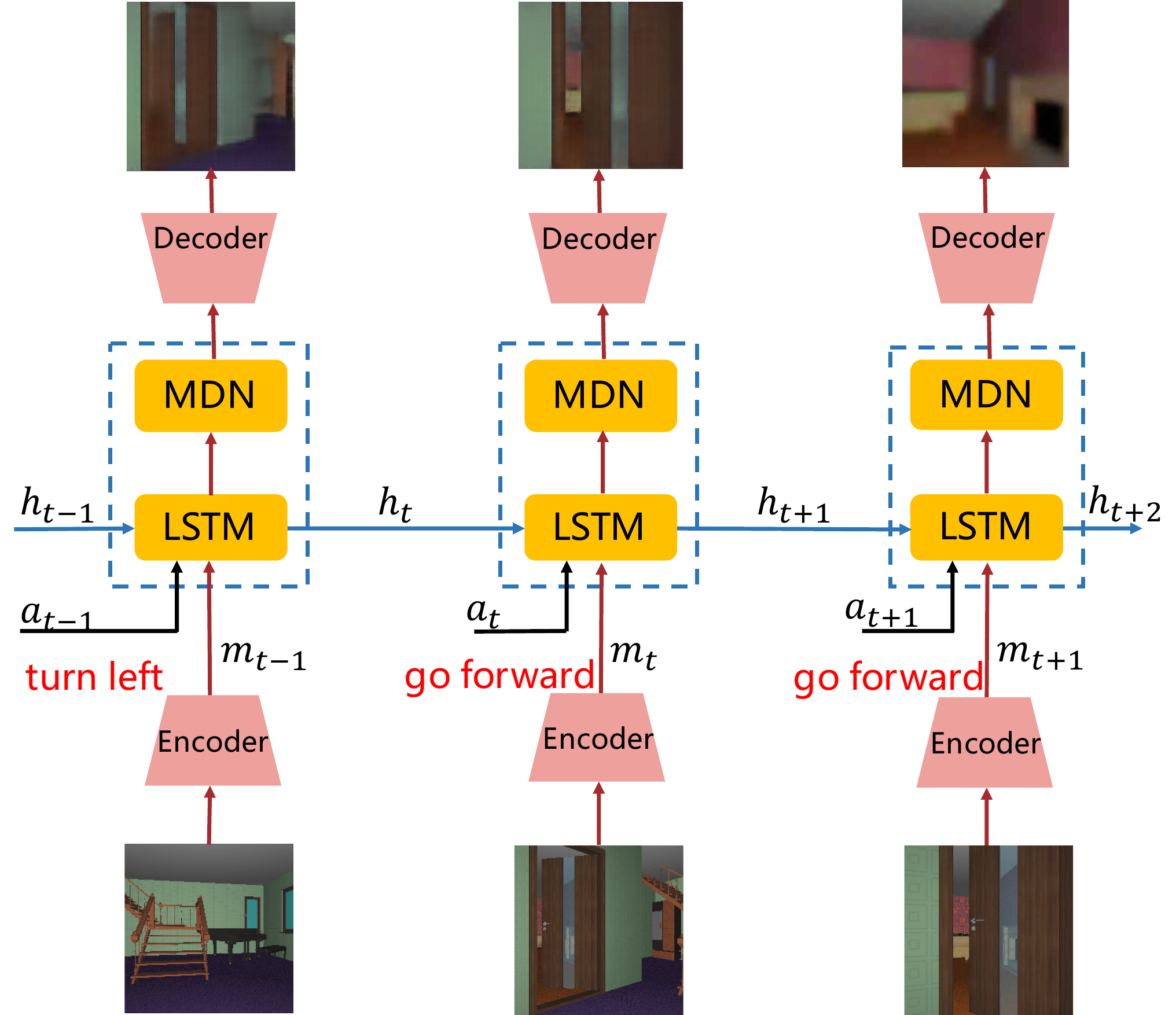}
		\caption{An example of the MIND Module. Mental Encoder outputs the mental representation of the current observation. Given the current mental representation and the next action, Imagery Model predicts the next mental representation. In this example, we use mental decoder to reconstruct the future observation which is the consequence of the given action. }
		\label{f2}
	\end{figure}

	Figure \ref{f2} showcases the internal process of the MIND module, which contains the mental encoder and the imagery model. Given the egocentric 224$\times$224 RGB image, our mental encoder first encodes it into mental representation $m_{t}$ and sends $m_{t}$  to our imagery model as input. The imagery model takes $m_{t}$, the previous hidden state $h_{t-1}$ and an action $a_{t}$ as inputs, and then predicts the next  $m_{t+1}^{'}$. For example, as shown in Figure \ref{f2}, at time t-1, we can know that the agent is facing a staircase through the observation $I_{t-1}$ it receives. The imagery model receives  `turn left' action, so it predicts the next mental representation $m_{t}^{'}$ before actually executing `turn left' action. The mental image $M_{t}$ at time t-1 is reconstructed by the mental decoder using $m_{t}^{'}$. It is similar to the observation $I_{t}$ at time t which is the real scene that the agent faces after turning left. From $M_{t}$, we can see that our MIND Module tries to imagine the scene that the agent will face after turning left. It means that without having to truly perform an action $a_{t}$, imagery model can predict the outcome which can help the agent to select a better action to perform. Moreover, the predictive mental images visualize the short-term goals of the agent, which make our method interpretable.
	\vspace{-0.3cm}
	
	\subsection{Training Procedure}
	In this subsection, we describe how to train our MIND module. Importantly, our MIND module is pretrained independently just using expert demonstrations in EQA dataset, without any additional annotation data to train it. In this task, The agent may be spawned at a random location in a 3D environment and may not immediately ‘see’ the scene containing the answer to the visual question. The expert demonstrations are trajectories following the shortest paths from the agent’s initial location to the target (more details in the experiment section). We first use these demonstrations to train our mental encoder to learn a mental representation of each frame and reconstruct the frame using $m_{t} $. We minimize the difference between the original frame $I_t$ and the reconstructed frame produced by the decoder from mental representation $m_{t}$. 
	
	After that, we can use our trained mental encoder and expert demonstrations to train our imagery model. Given mental representation $m_{t}$ encoded by mental encoder and the action $a_{t}$ that agent performed, imagery model predicts the next  $m_{t+1}'$. We minimize the difference between $m_{t+1}'$ and the real $m_{t+1}$. Notice that $m_{t+1}'$ is not corresponding to the next frame in trajectory after executing an atomic action. We expect our MIND module to predict a further outcome of several actions instead of just an atomic action, which is more helpful for generating imaginary subgoals.
	
	\section{Walking with MIND } \label{s4}
	After pre-training the MIND module, we can apply it to navigating. Our navigation model is based on the PACMAN \cite{Das_2018_CVPR}, which is a hierarchical model that decomposes the navigator into a planner and a controller.
	\begin{figure*}[h]
		\centering
		\includegraphics[width=\textwidth]{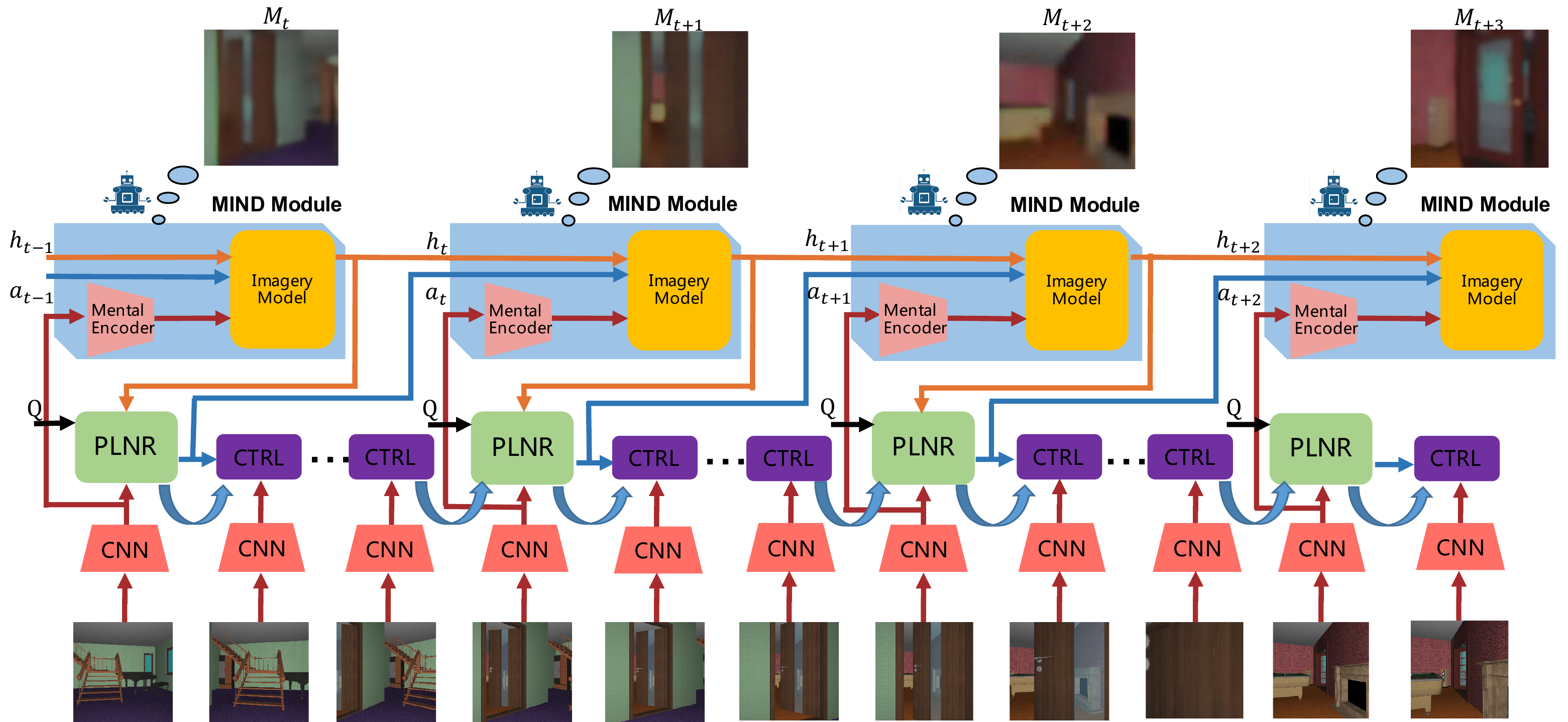}
		\caption{The overview of the MIND agent. Before deciding an action to execute, it predicts some short-term subgoals in the mind, which yields better planning efficiency and visualizes the agent’s intentions in a way that human can understand. The images above the MIND Module are the mental images generated by the MIND module.}
		\label{f3}
		\vspace{-0.2cm}
	\end{figure*}
	
	\subsection{PACMAN Navigator}
	PACMAN navigator contains a planner and a controller. The planner selects actions(\textsl{i.e. forward, turn-left, turn-right, stop}) and the controller decides how many times to perform the primitive action. More specially, the planner first selects an action and gives control to the controller. The controller outputs 0 or 1 that 0 means to stop and return control to the planner, and 1 means to execute the action that the planner has chosen once. Besides, the controller must return control to the planner after five consecutive 1. One forward action is equivalent to 0. 25 meters and one turn right or turns left action is equivalent to $9^{\circ}$ change in viewing angle. The planner is instantiated as an LSTM, and the controller is instantiated as a multilayer perceptron. Formally, the planner produces an action $a_{t}$ as follows:
	\begin{equation}
	a_{t}, h_{t} \leftarrow PLNR(h_{t-1}, I_{t-1}^{0}, Q, a_{t-1})
	\end{equation}
	where $Q$ is the encoding of the question, and $I_{t}^{n}$ is the encoding of the observed image at t-th planner-time and n-th controller-time. 
	
	The controller produces 0 or 1 as follows:
	\begin{equation}
	\{0, 1\} \leftarrow CTRL(h_{t}, I_{t}^{n}, Q, a_{t})
	\end{equation}		
	where 0 means to stop and return control to the planner, and 1 means to execute the action that the planner has chosen once. 
	
	\subsection{Plan in MIND}
	In our approach, the planner takes the imagery model's hidden state $h_{t}$ as an extra input and selects an action to perform. The MIND module executes at each Planner timestep to help Planner choose better action. At controller-time, it is the same as PACMAN model. Specifically, at each planner-time step, the planner selects an action based on the question encoding, current observation, and its hidden state. Further, it takes into consideration the mental images generated by the MIND module. As illustrated in Figure \ref{f3}, the mental images above the MIND module are the imaginary consequence of performing a sequence of specific actions, and they visualize the short-term subgoals of our agent. For example, at time t+3, our agent has just entered the room, and the mental image $M_{t+3}$ predicts the consequence of performing several forward actions. The Mental image $M_{t+3}$ shows the imaginary scene of the room's front door, which indicates that the short-term goal of our agent at time t+3 is to go to the front door of the room.
	
	With the mental imagery enhanced module, our agent can generate imaginary short-term subgoals that related to the final objectives, which improves its planning efficiency. Also, the MIND module introduces a basic understanding of the environment to the agent, which enhances its generalizability to unseen buildings and guarantees its performance when training data is few. For example, with such knowledge, our agent may know that it can only leave the room through the door, so it may not hit the wall many times to learn how to get out of the room. Therefore, we combine the planner in PACMAN with the MIND module. 
	
	\subsection{Reinforcement Learning of the MIND Agent}
	In this subsection, we will describe how to train the MIND module and the navigator together in details. We first use imitation learning to warm-start our MIND agent and then use A3C\cite{mnih2016asynchronous} for fine-tunning.\\
	\textbf{Imitation Learning to Warm-start:  }In EQA dataset, there are four kinds of questions (more details will be described in the next section) and each question has a target object. The question is about properties (\textsl{e.g. location, color}) of the target object (\textsl{e. g. `What color is the sofa?'}), so it enables to generate expert demonstrations for imitation learning using the shortest path from the agent's initial position to the target object. We use these expert demonstrations to warm-start our agent. We train our agent to mimic the expert demonstration using behavior cloning. More specially, given the current observation, question, our agent is trained to select the right action on the expert demonstrations. We find it hard to let it learn expert demonstrations directly because the initial position in expert demonstrations is too far from the target object. Similar to Das's \textsl{et al.} work \cite{Das_2018_CVPR}, we using distance-based curriculum learning to train our model. Firstly, we initialize our agent five steps away from the target object along the expert demonstrations and let it mimic the remaining actions in the expert trajectories. After it learns the remaining actions successfully, we backtrack five additional steps. Finally, our agent can mimic expert demonstrations after 20 epochs. The cost function $J_{bc}(\pi_{\theta})$ can be written as:
	\begin{equation}
	J_{bc} = E_{a \sim \pi_{\theta}}[log(\pi(a_{t}|s_{t}))]
	\end{equation}
	where $a_{t}$ is the demonstration action and $s_{t}$ is the state, containing the current frame, mental image, question encoding, and the navigation history. The training objective is to maximize this function.\\
	\textbf{Reinforcement Learning to Fine-tune:  }After behavior cloning, we use reinforcement learning to endow the agent with the ability to recover from wrong actions and encourage our MIND module to generate more task-related mental imagery.
	
	Inspired by \cite{Das_2018_CVPR}, we propose an actor-critic RL framework based on three kinds of rewards, namely \textbf{final reward} $r_{f}$, \textbf{progressive reward} $r_{p}$ and \textbf{planned reward} $r_{m}$ that encourage our agent to reach the target location efficiently and give a correct answer. Importantly, the planned reward $r_{m}$ encourage our agent learning to form more task-related and objective-related short-term subgoals.
	
	The final goal of our agent is to answer the questions correctly. Therefore, we define the final reward to reflect whether the question is answered correctly. We use the same question-answering model as \cite{Das_2018_CVPR}. Question-answering model is called when our agent chooses to stop. It receives the question, and the image features from the last five frames along the navigation path and then computes image-question similarity for each image between the question encoding and the image features. These similarities are used as attention weights to combine these five image features with the question encoding. The final question-image features are passed through a softmax classiﬁer to predict a distribution over 172 answers. Let T denote the last time step. The final reward is defined as:
	\begin{equation}
	r_{f}(s_{T}, a_{T})=\begin{cases}
	1 + \lambda_f \cdot max(N_{max}-n, 0 ) & \text{if answer is correct}\\
	0& \text{else}
	\end{cases}
	\end{equation}

	Different from Das's \textsl{et al.} work, we set final reward as one plus the weighted maximum between 
	the maximum number of actions $N_{max}$ minus the actual number of actions n that our agent executes and zero if the answer is correct, and zero else. The second item in the final reward is aimed to encourage our agent to perform fewer actions to answer the question, which can improve the efficiency of navigation. We can adjust weight parameter $\lambda_f$ to balance between the correct answer and the navigation efficiency.
	
	The progressive reward is an intermediate reward that encourages our agent get close to the target object. Let $Dist(s_{t})$ denotes the distance between the location at state $s_{t}$ and the target location. Then the progressive reward after taking action $a_{t}$ at state $s_{t}$ is defined as:
	\begin{equation}
	r_{p}(s_{t}, a_{t}) = Dist(s_{t}) - Dist(s_{t+1})
	\end{equation}
	If the distance to the target location becomes smaller after taking action $a_{t}$ at state $s_{t}$, then our agent will get a positive reward that reflects how much the distance from the target has been reduced. Also, if our agent goes further with the target, it will be punished by a negative reward. 
	
	The progressive reward only considers the current effect but ignores the impact on the future. For example, in order to get closer to the target, it may go to the corner of a room instead of leaving the room, which may temporarily reduce the distance to the target location. However, it can never get to the target location without leaving the room. 
	
	To account for this, we define an intermediate reward called \textsl{planned reward}. As mentioned above, the mental images produced by the MIND module reflect the imagery short-term subgoals in our agent's mind, and we can use them to inspect whether the agent's next few actions are beneficial to the final objective. Therefore, we define the planned reward as the improvement of the correct answer's probability. More specifically, let $I_{t-1}^{-3}$ denote the third last frame at $(t-1)$-th planner-time and $M_t$ denote the mental image at $t$-th planner-time. Let $P_a(o^{*}|...)$ denote the probability of the correct answer produced by the question-answering model, and $o^{*}$ is the correct answer among 172 candidates. The planned reward is written as:
	\begin{equation}
	r_{m}(s_{t}, a_{t}) = P_a(o^{*}|I_{t-1}^{-3}, ... , I_{t}^{0}, M_{t}) - P_a(o^{*}|I_{t-1}^{-4}, ... , I_{t}^{0})
	\end{equation}
	At each planner-time step, we compute the probability of the correct answer based on the current mental image $M_{t}$ and the last four frames $I_{t}^{0}, ... , I_{t-1}^{-4}$, and the probability based on the last five frames. We compare them to inspect whether the current subgoal of the agent is beneficial to answer the question. With the mental image, if the probability of the correct answer increases, it means that our MIND agent forms a task-related and objective-related short-term subgoals. If not, the agent will be punished by a negative reward. At test time, we just call answer model once when our agent stops.
	
	So far, we have 3 kinds of rewards. They inspire an efficient navigation path to the target location and a correct answer to the given natural language question. We add three rewards together as the total reward function:
	\begin{equation}
	R(s_t, a_t)=\begin{cases}
	r_{p}(s_{t}, a_{t}) + r_{m}(s_{t}, a_{t}) + r_{f}(s_t, a_t) & \text{if t = T}\\
	r_{p}(s_{t}, a_{t}) + r_{m}(s_{t}, a_{t}) & \text{else}
	\end{cases}
	\end{equation}
	Then, we use A3C \cite{mnih2016asynchronous} with generalized advantage estimator (GAE)  \cite{schulman2015high} to optimize our policy. The gradient of $J_{rl}$ can be written as:
	\begin{equation}
	\nabla_{\theta} J_{rl} = E_{a \sim \pi_{\theta}} [ \nabla_{\theta}[ -(\sum_{l}^{T}(\gamma \lambda)^{l}\cdot\delta_{t+l}^V)\cdot log(  \pi_{\theta}(a_{t}|s_{t}) ) + V_{loss}(s_{t})] ]
	\end{equation}
	
	\begin{equation}
	V_{loss}(s_{t}) = (V_\theta(s_t) - R(s_t, a_t))^2
	\end{equation}
	
	\begin{equation}
	\delta_{t+l}^V = -V(s_{t + l}) + \gamma \cdot V(s_{t + l + 1}) + R(s_{t+l})
	\end{equation}
	where $\delta_{t+l}^V$ is the generalized advantage estimator and $V_{\theta}(s_{t})$ is the estimated value of state $s_{t}$ produced by the critic for $\pi_{\theta}$. 
	
	\section{EXPERIMENTS} \label{s5}
	
	
	\subsection{EQA  Dataset}
	\textbf{Statistics:  }EQA dataset contains about 9000 questions in 774 environments. More specially, there are 7129 training data in 648 environments, 853 validation data in 68 environments and 905 testing data in 58 environments and there are no overlapping environments between them. Thus, the performance on test set directly reflects the generalizability to novel novel environment.\\
	\textbf{Question Form:  }In EQA dataset, there are 4 kinds of questions as shown below:
	\begin{itemize}
		\item location: ‘What room is the <OBJ> located in ?’
		\item color: ‘What color is the <OBJ>?’
		\item color\_room: ‘What color is the <OBJ> in the <ROOM>?’
		\item preposition: ‘What is <on/above/below/next-to> the <OBJ> in the <ROOM>?’
	\end{itemize}
	There are 10 kinds of <ROOM>, such as dining room, bathroom and bedroom. There are 50 kinds of <OBJ>, such as bed, table and coffee machine. 

	\subsection{Evaluation Metric}
	In order to answer the question, the agent must move from a random initial position to the target location that contains the answer to the visual question. For instance, to answer `What room is the shoe rack located in?', the agent must perceive the environment and perform a sequence of correct actions to move to the room with the shoe rack. In EQA dataset, the target location is marked by humans. Let $d_{0}$ denotes the initial distance to the target, $d_{T}$ denotes the final distance (how far is the agent from the goal when it stops) to the target and $d_{\Delta}=d_{0} - d_{T}$ denotes the change. We use $d_{\Delta}$ to evaluate the navigation performance and spawn agent 10, 30, or 50 primitive actions away from target, denoted as $T_{-10}, T_{-30}, T_{-50}$. The bigger $d_{\Delta}$ indicates the agent has more ability to find the target location containing the visual answer. We use accuracy to evaluate the answering performance.

	\subsection{Setup}
	The laten space dimension of $\beta$-VAE is 128. When training Mixture Density Network, it is easy to meet a numerically unstable problem. To avoid it, we use the log-sum-exp trick and gradient clipping technique and replace exponential function to ELU(1,x)+1. We have explored several different structures of the imagery model, and the best version we have is 1 LSTM layer, 5 Gaussians, 512 hidden units. We use Adam optimizer with a learning rate of 1e-4 to train mental encoder and with a learning rate of 1e-5 to train imagery model. We set the maximum number of actions $N_{max} = 80$, and the batch size is 20. During A3C fine-tuning, we set $\gamma = 0.99$,  $\lambda = 1.00$ and the learning rate is 1e-4.
	
	\begin{table}[]
		\caption{Evaluation of EmbodiedQA agents on navigation and answering metrics for the EQA test set.}
		\label{t1}
		\resizebox{\linewidth}{!}{
			\begin{threeparttable}
				\begin{tabular}{ lcccccc}
					\toprule
					&\multicolumn{3}{c}{Navigation$(d_{\Delta})$}  &\multicolumn{3}{c}{QA(accuracy)}     \\
					&$T_{-10}$ & $T_{-30}$ & $T_{-50}$      &$T_{-10}$ & $T_{-30}$ & $T_{-50}$  \\
					
					\midrule
					PACMAN(BC)                           & -0.04   & 0.62   & 1.52  &	48.48\%	 &	40.59\%	    &	  39.87\%         \\
					PACMAN(BC+REINFORCE)       & 0.10     &  0.65  & 1.51    &	50.21\% & 42.26\%      &     40.76\%        \\
					NMC(BC)\tnote{**}                                 & -0.29   &  0.73  & 1.21     &	43.14\%  & 41.96\%      &      38.74\%      \\
					NMC(BC+A3C)\tnote{**}                        & 0.09     &  1.15   & \textbf{1.70}  &	53.58\% & 46.21\%      &   44.32\%    \\
					\midrule
					Blindfold                                &-0.02  & -0.13  & -0.44  & 50.34\%   & \textbf{50.34\%} & \textbf{50.34\%} \\
					\midrule
					MIND(BC)                                         & 0.17     &  0.94   & 1.52    &	 50.34\%	 &	40.02\%	     &	   39.13\%            \\       
					MIND(BC+A3C)     & \textbf{0.25}     &  \textbf{1.21}   & 1.65 &	  \textbf{54.83\%}	  &	\textbf{46.71\%}	   & \textbf{44.56\%}  \\
					\bottomrule
				\end{tabular}
				
				\begin{tablenotes}
					\footnotesize
					\item[**] this approach requires additional training data annotated with a sequence of subgoals, which is not available in EQA dataset.
				\end{tablenotes}
			\end{threeparttable}
		}
	\vspace{-0.4cm}
	\end{table}
	
	\subsection{Results \& Ablation Analysis}
	\textbf{Overall Analysis: }  We compare our MIND agent with PACMAN\cite{Das_2018_CVPR}, NMC \cite{das2018neural}, and Blindfold \cite{anand2018blindfold}. Blindfold is a question-only BoW baseline. Although it achieves the best QA accuracy by leveraging the biases in the dataset, its navigation accuracy is poor. 
	As shown in Tab \ref{t1}, our MIND(BC+A3C) achieves better $d_{\Delta}$ at $T_{-10}, T_{-30}$ and better QA accuracy at all distance compared with PACMAN and NMC. This suggests that our MIND agent has stronger generalizability. This gain mainly comes from the MIND module's ability to plan short-term goals instead of just executing primitive actions over long time horizons and modelling the dynamics of the environment. Even if MIND(BC) isn't fine-tuned using A3C, it performs better than PACMAN(BC + Reinforcement) in $d_{\Delta}$. This fact proves the effectiveness of our MIND module. Comparing MIND(BC) with MIND(BC + A3C), we can see that our RL framework signiﬁcantly boosts performance in answering accuracy. NMC achieves best navigation performance at $T_{-50}$, since it has more annotated well-desighed subgoals(e.g., find a specific object or room, exit a specific room) which are crucial for long-term planning. \\
	\begin{figure}[]
		\centering
		\subfigure[Navigation]{
			\includegraphics[width=0.45\linewidth]{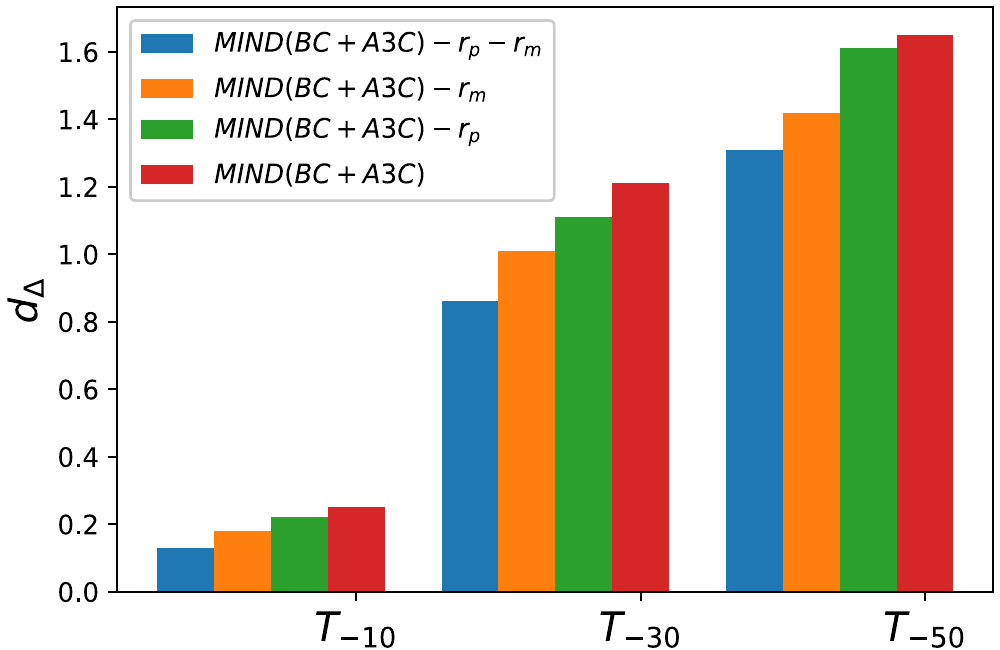}
		}
		\quad
		\subfigure[Answering]{
			\includegraphics[width=0.45\linewidth]{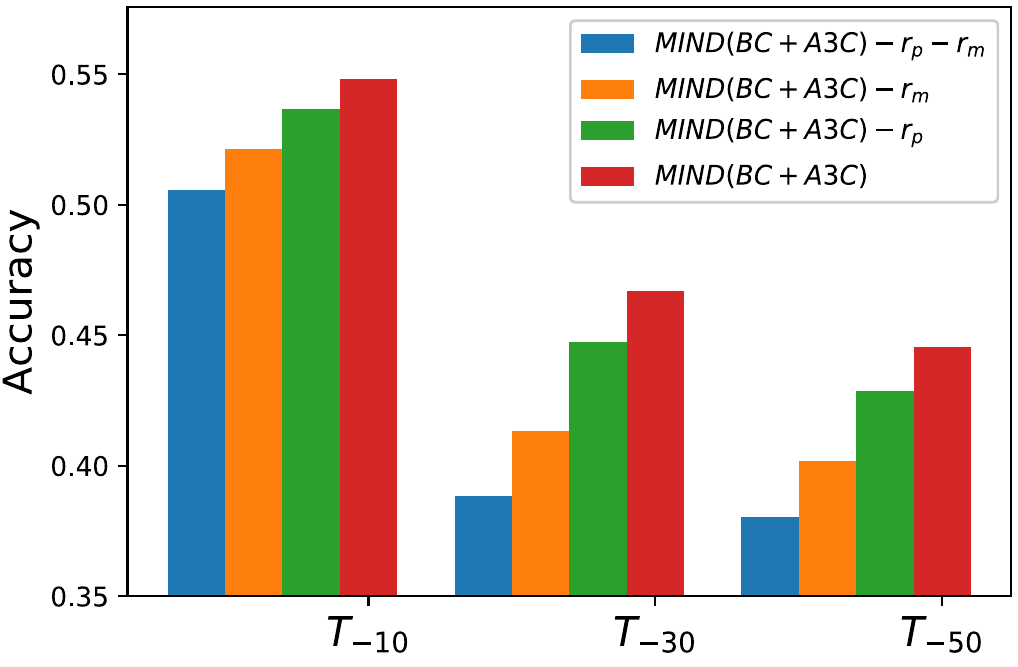}
		}
		\caption{Results of MIND agent with different reward. MIND(BC+A3C)$-r_*$ means that we train the agent without $r_*$}
		\label{f5}
		\vspace{-0.5cm}
	\end{figure}
	\begin{figure}[htpb]
		\centering
		\includegraphics[width=\linewidth]{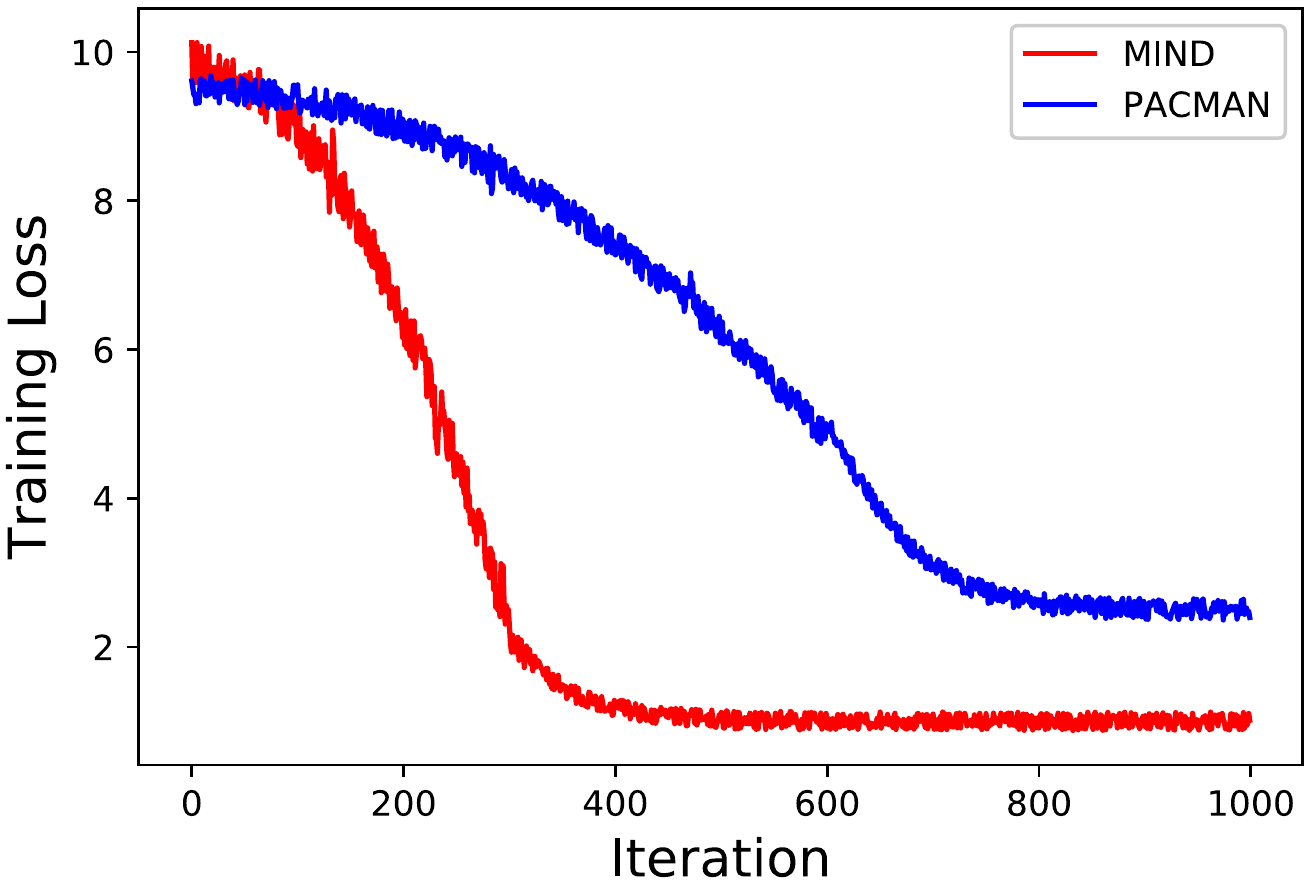}
		\caption{Learning curves of MIND and PACMAN agent.}
		\label{f6}
		\vspace{-0.4cm}
	\end{figure}
	\textbf{Eﬀect of Diﬀerent Rewards: }To fully investigate the effectiveness of the three different rewards, we conduct an ablation analysis on these rewards. We only use $r_f, r_f + r_p$ and $r_f + r_m$ to train thress agents separately, and compare them with our best model. MIND(BC+A3C)$-r_*$ means that we train the agent without $r_*$. As shown in Figure \ref{f5}, without intermediate rewards $r_p$ and $r_m$, MIND(BC+A3C) $-r_p - r_m$ performs even worse than MIND(BC), which reflects the original A3C tranning can't improve performance. Comparing MIND(BC+A3C) $-r_m$  with MIND(BC+A3C) $-r_p$, we find that the gain from planned reward $r_m$ is higher than progressive reward $r_p$, it is because the planned reward encourages our MIND module to generate more task-related imagery, which enhances the navigational performance.\\
		\begin{figure}[htpb]
		\centering
		\includegraphics[width=\linewidth]{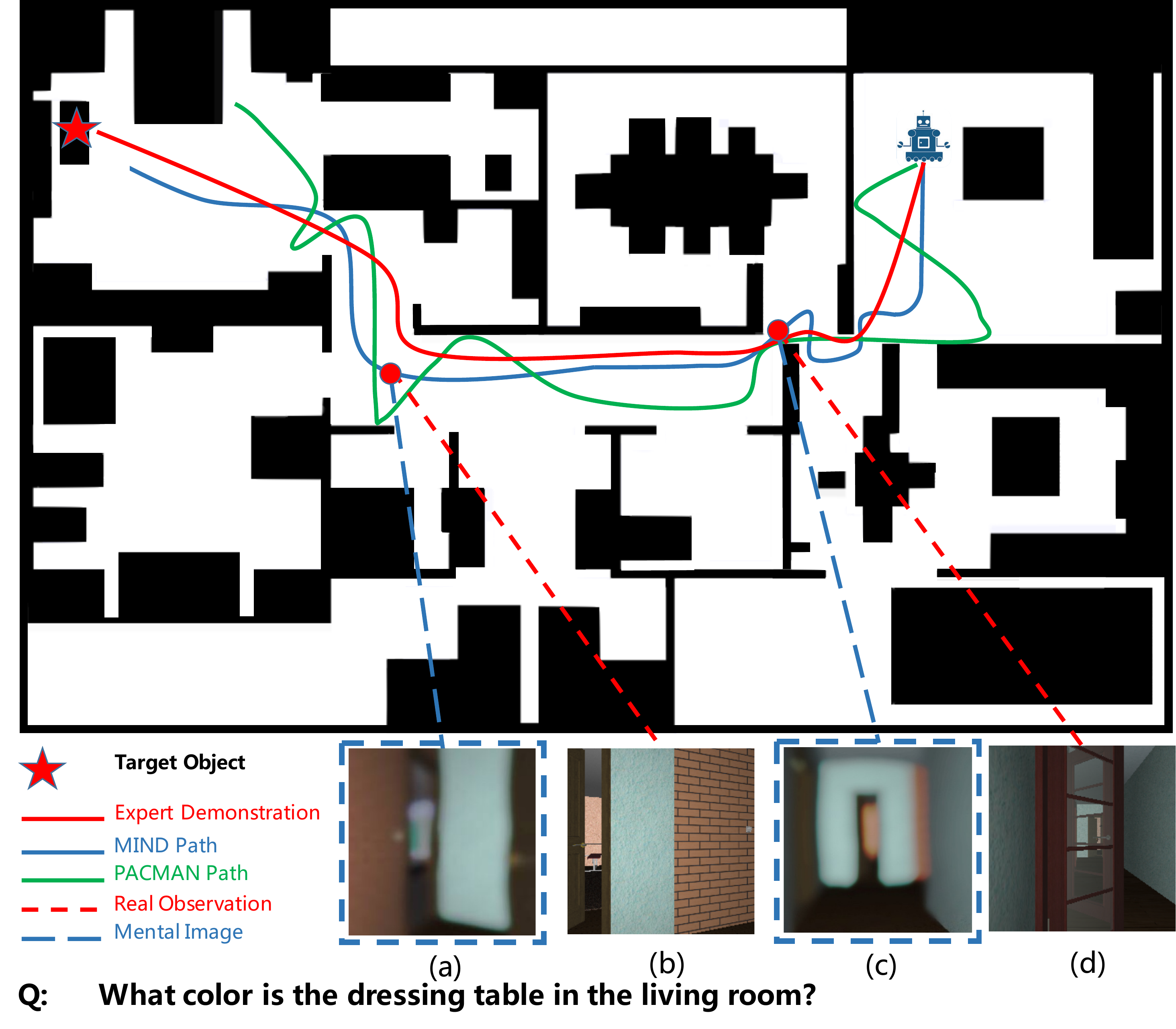}
		\caption{Example trajectories executed by PACMAN, MIND and the human expert. The global trajectories are shown in a top-down view (the top-down view is not available to the agents). The black areas represent obstacles, which can not be directly passed by the agents. The red trajectory is the human demonstration. The blue trajectory and green trajectory are executed by MIND and PACMAN, respectively.}
		\label{f7}
		\vspace{-0.4cm}
	\end{figure}
	\begin{figure}[]
		\centering
		\subfigure[]{
			\includegraphics[width=0.45\linewidth]{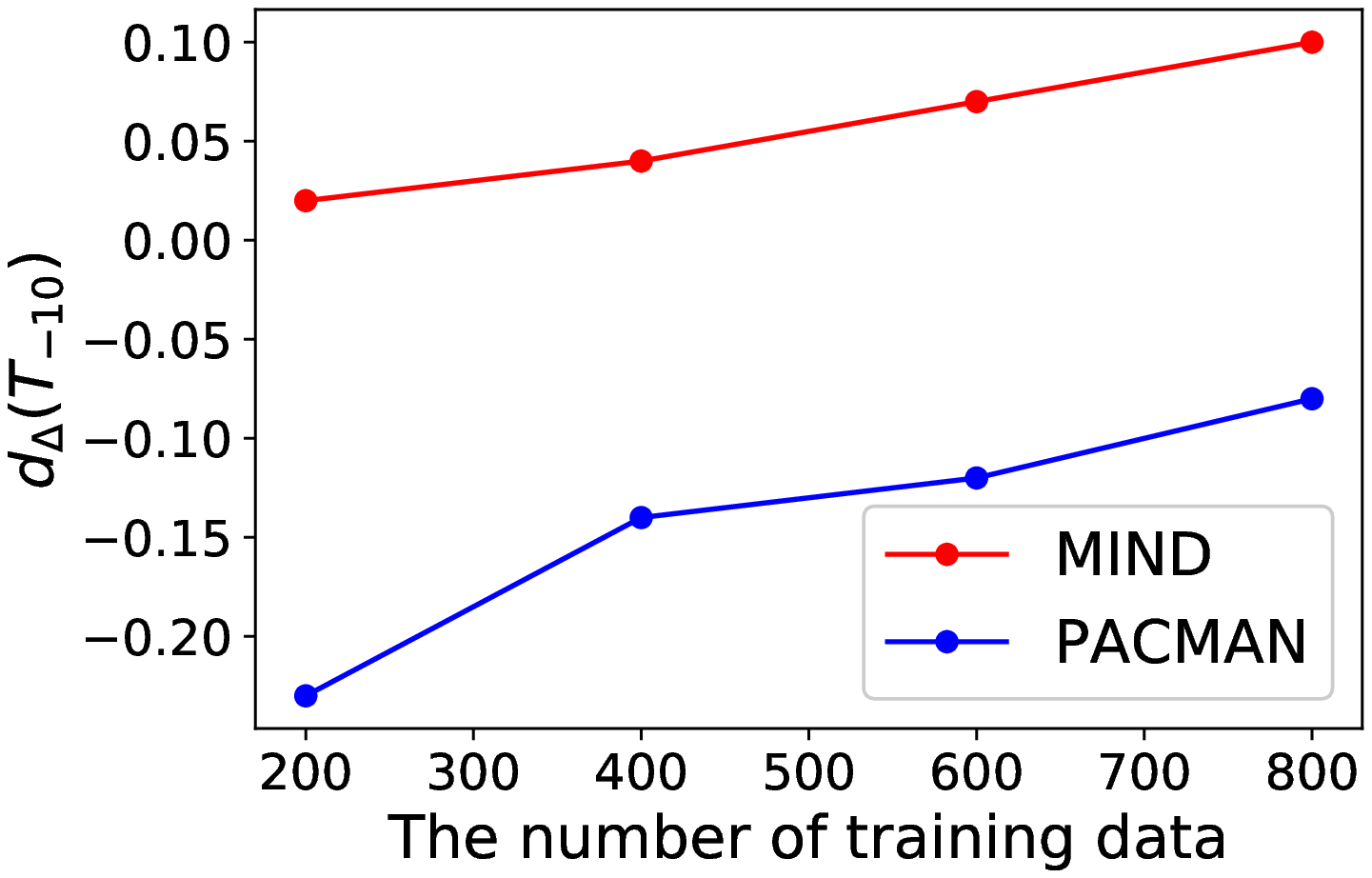}
		}
		\quad
		\subfigure[]{
			\includegraphics[width=0.45\linewidth]{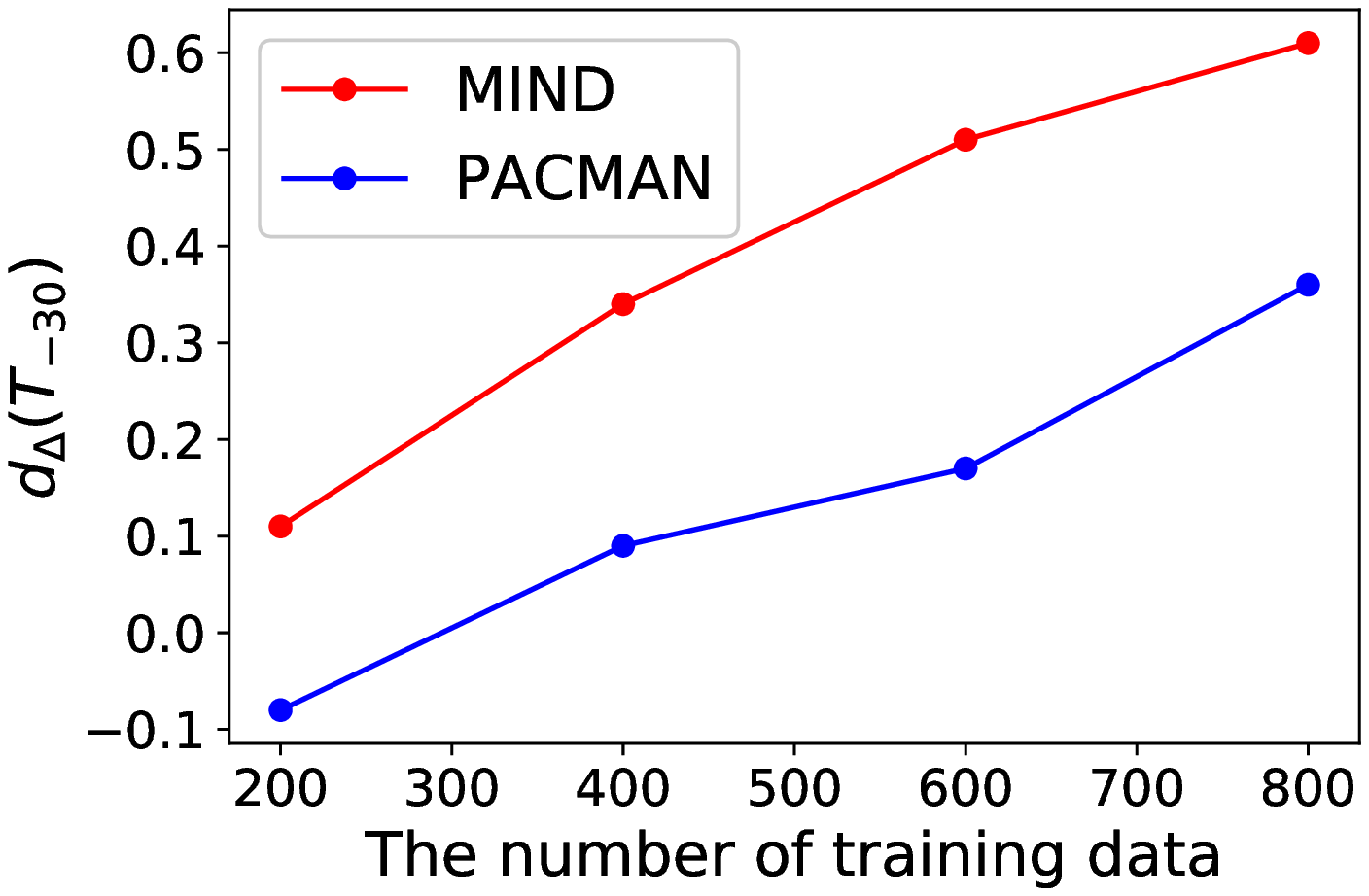}
		}
		\quad
		\subfigure[]{
			\includegraphics[width=0.45\linewidth]{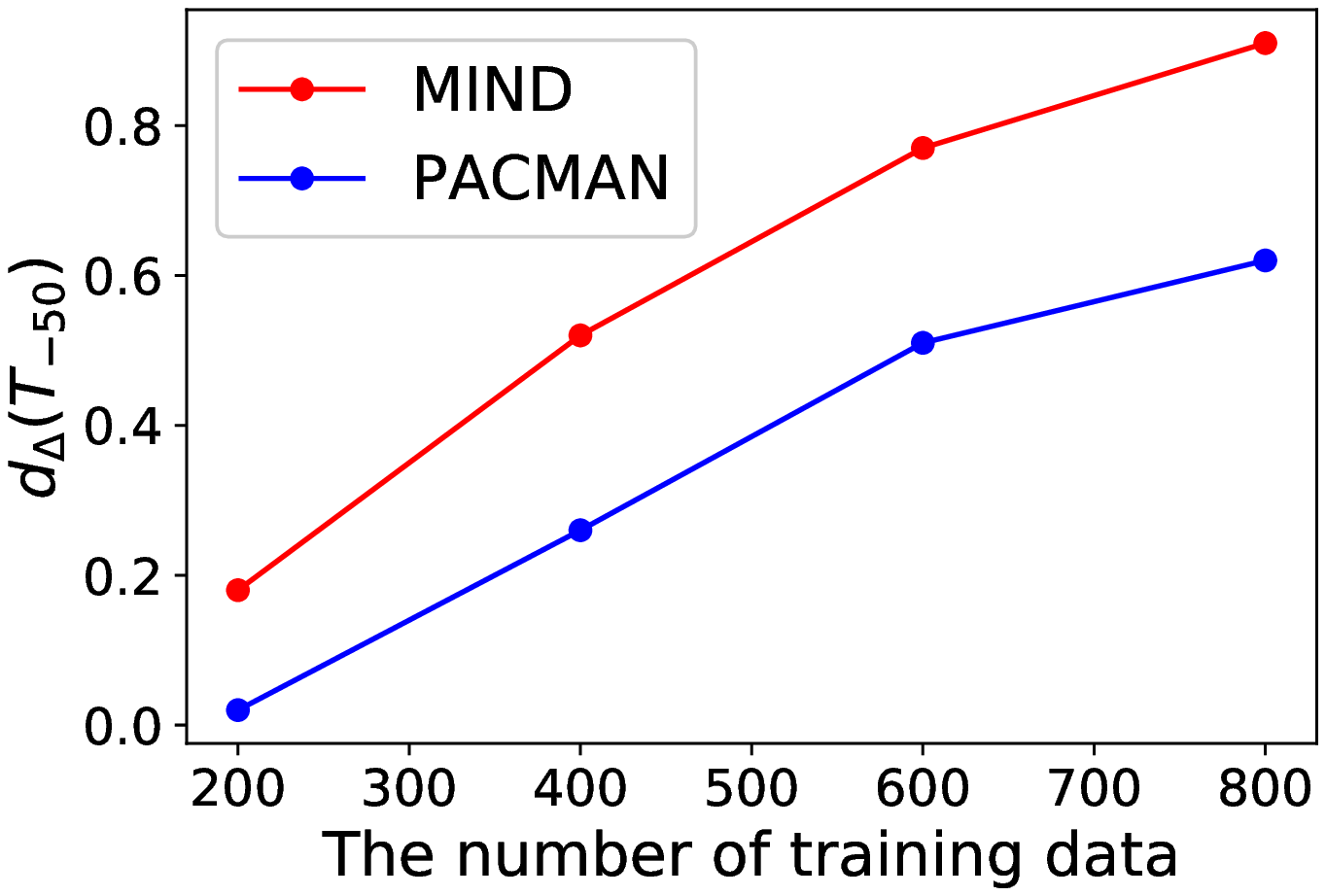}
		}
		\quad
		\subfigure[]{
			\includegraphics[width=0.45\linewidth]{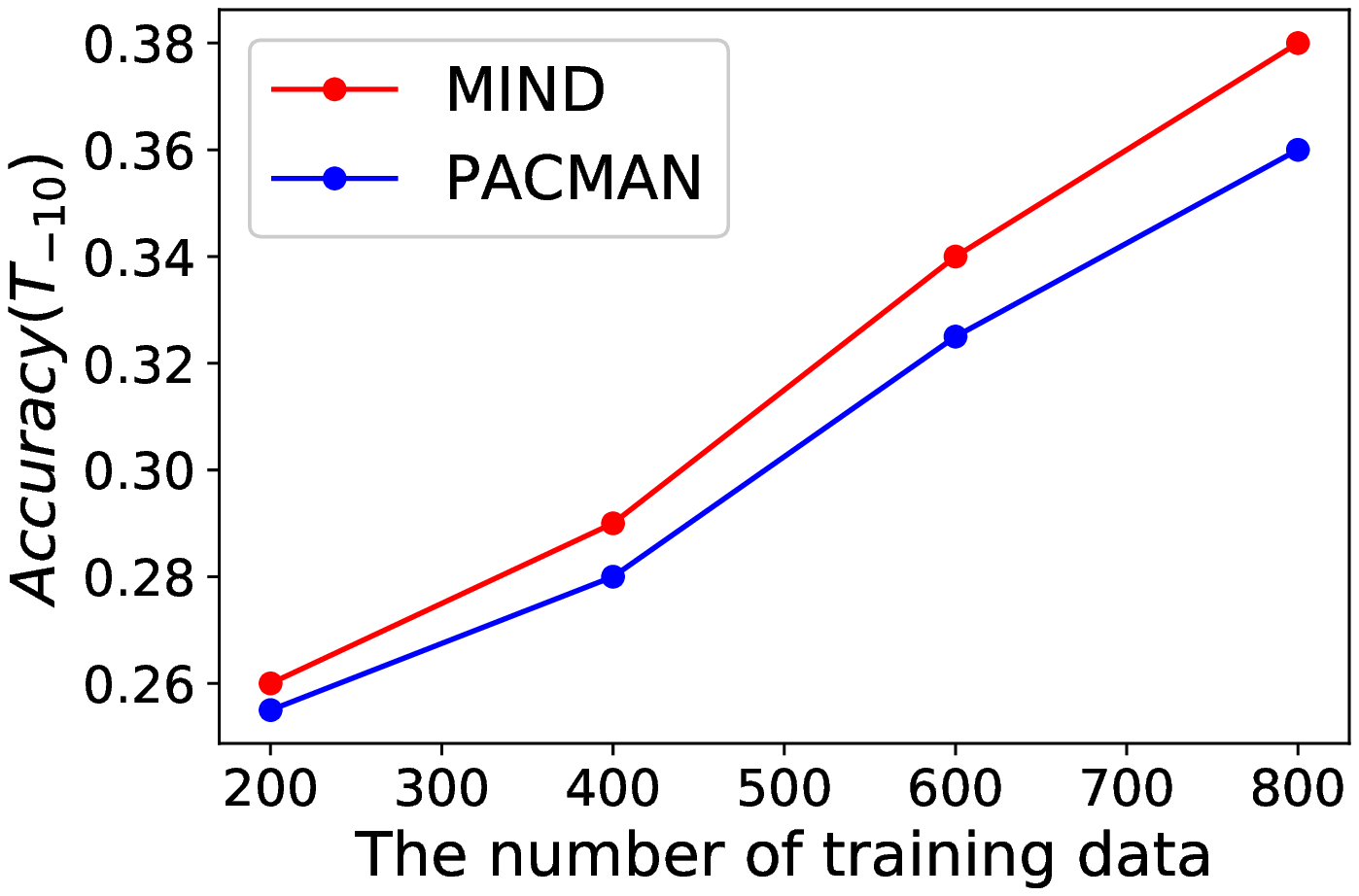}
		}
		\quad
		\subfigure[]{
			\includegraphics[width=0.45\linewidth]{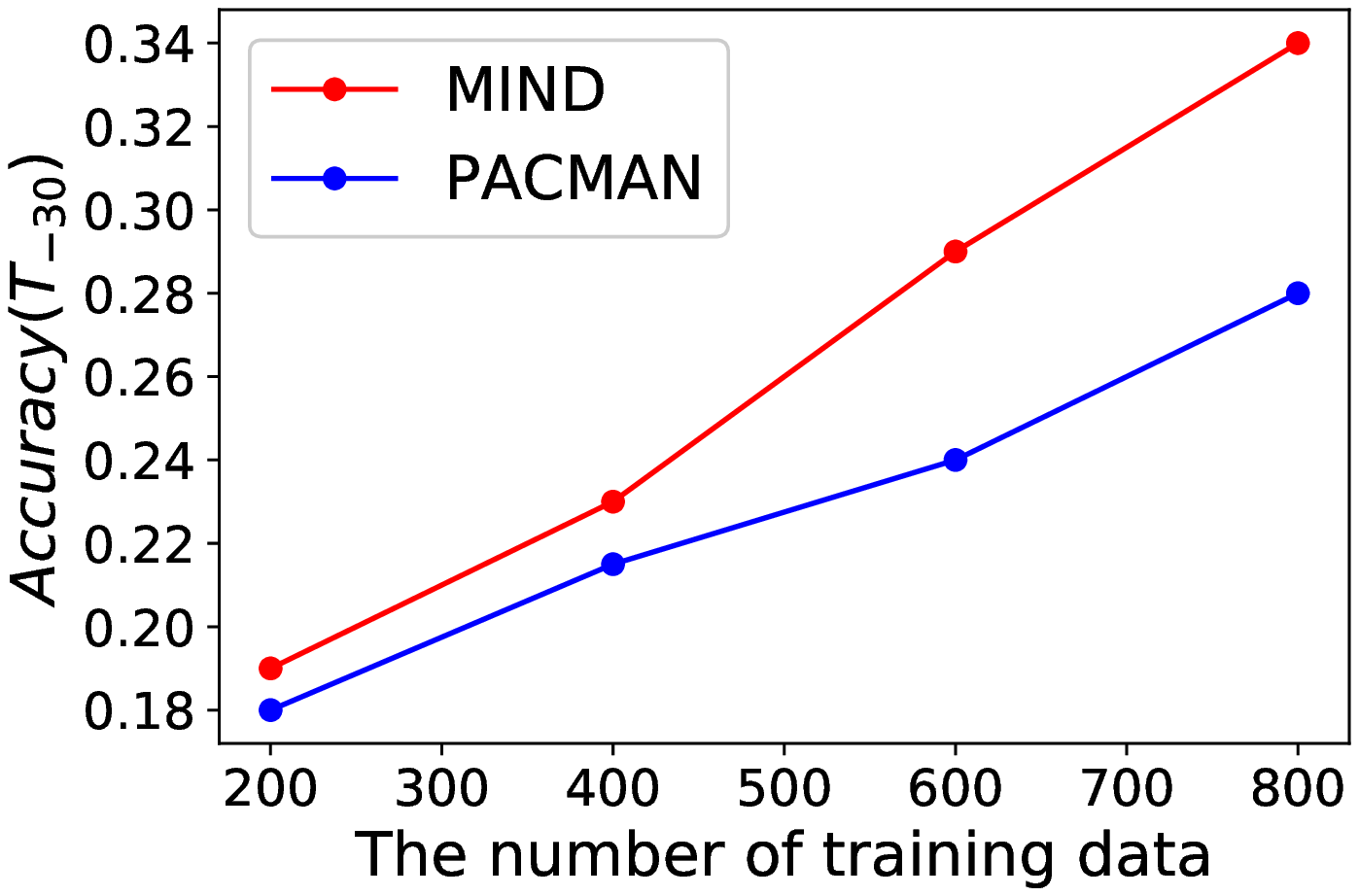}
		}
		\quad
		\subfigure[]{
			\includegraphics[width=0.45\linewidth]{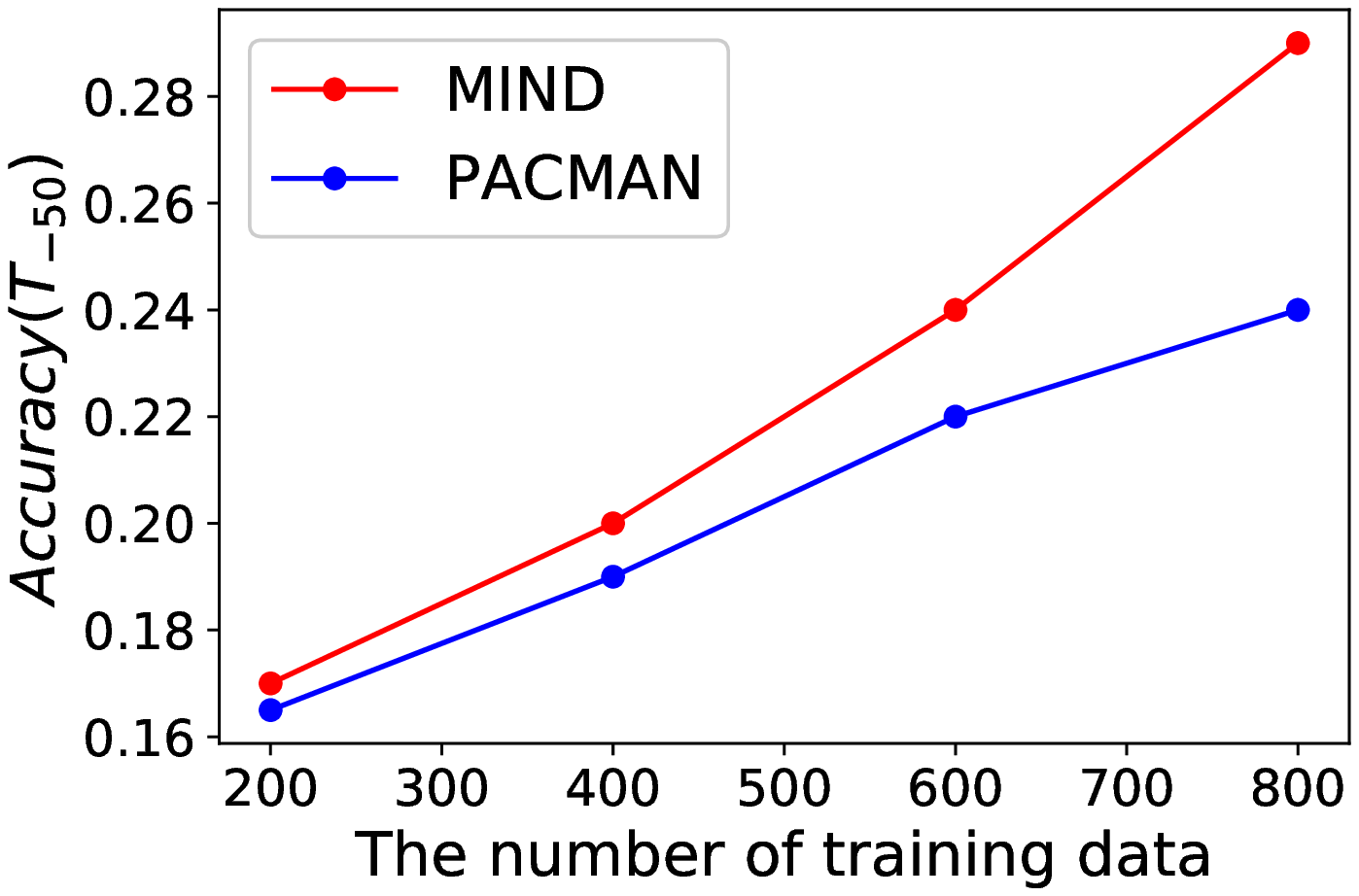}
		}
		\caption{The navigation performance $d_{\Delta}$ and answering accuracy at $T_{-10}, T_{-30}, T_{-50}$. }
		\label{f4}
	\end{figure}
	\textbf{Generalizability \& Convergence Speed: }To evaluate the ability to generalize from a few samples, we use part of validation data to train MIND agent and PACMAN, and then compare the generalizability of them on the testing data. The size of validation (853 questions in 68 environments) is only about one-tenth of the training (7129 questions in 648 environments). Our MIND module is pretrained independently using training data, and it has never `seen' any environment in validation data or testing data. There are no overlapping environments between them, so this experiment can strictly test agent's generalizability to unseen environments with a few training data. We use a different number of validation data as our training data in this experiment to train agents, and compare their performance on navigation and question answering. For a fair comparison, the PACMAN and MIND are both trained using behavior cloning to warm-start and REINFORCE \cite{williams1992simple} to fine-tune. We also show their learning curves when we train them using REINFORCE.
	
	As shown in Figure \ref{f4}, at each size of training data, MIND agent performs better than PACMAN both in navigation performance and answering accuracy. We can see that PACMAN almost does not work when the number of training data is 200. From Figure \ref{f6}, we can see that our MIND agent converges faster and better.\\
	\textbf{Planning Efficiency \& Behavioral Interpretation: } To demonstrate our method's superior performance on route planning and behavioral interpretation, we carry out a case study. As shown in Figure \ref{f7}, 
	the agent is spawned in the kitchen at the top right of the top-down view and asked a question. We can see that PACMAN agent first goes to the left. Although its direction is correct, it can not walk out of the kitchen through the wall, and it needs to try several times to get out of the room. It is because that it lacks the basic understanding of the environment (\textsl{e.g. `it can only leave the room through the door'}) and the short-term planning (\textsl{e.g. `it first need to get out of the kitchen'}). It just wants to get close to the target object, but it does not know it has to get out of the kitchen first. Therefore, it has low navigation efficiency. In contrast, our MIND agent acts more like humans. It plans short-term goals in mind, performs a sequence of actions to achieve them, and finally reaches the target location and answers the question. From the image (d) in Figure \ref{f5}, we can see that the MIND agent is entering the corridor. The mental image (c) visualize its short-term subgoal that it intends to go straight and get to the end of the corridor, which is proved correct by MIND's trajectory in a top-down view. When it gets to the end of the corridor, we can see that it plans to turn right and get close to the bedroom, which is shown in mental image (a). These two cases suggest that the MIND agent holds mental images in its mind, which are the short-term goals of itself and make its actions more interpretable and planned. \\
	\textbf{Further Discussion: } Recently, we noticed that Wu \textsl{et al.}\cite{wu2019revisiting} proposed a simple baseline that can be end-to-end trained, which is competitive to the state-of-the-art. Their empirical results indicate that the QA bottleneck is due to the worse navigation ability, and current approaches are far from satisfaction. Further, Wu \textsl{et al.} introduced an easier and practical setting for EmbodiedQA. They propose a proxy task for the agent to explore the new environment by randomly placing some makers, which helps the agent to adapt the learned model to the new environment. This practical setting can be well applied to other approaches, and improve their generalizability to the new scene. Also, the text-only baseline\cite{anand2018blindfold} inspires us to create less biased QA pairs in the future.
	\vspace{-0.3cm}
	\section{Conclusion} \label{s6}
	In this paper, we propose the \underline{M}ental \underline{I}magery e\underline{N}hance\underline{D} Module for EmbodiedQA and a Deep Reinforcement Learning framework for MIND agent. The MIND module models the environment dynamics and predicts mental images that relate to the final goal, which endows our agents with strong generalizability and interpretability; and improves its planning efficiency. 
	The experimental results and further analysis prove that the agent with the MIND module is superior to its counterparts not only in EQA performance but in many other aspects such as route planning, behavioral interpretation, and the ability to generalize from a few examples.  
	 
	\vspace{-0.3cm}
	\section*{Acknowledgment}
	This work has been supported in part by National Key Research and Development Program of China (SQ2018AAA010010), NSFC (No.61751209, U1611461), Hikvision-Zhejiang University Joint Research Center, Zhejiang University-Tongdun Technology Joint Laboratory of Artificial Intelligence, Zhejiang University iFLYTEK Joint Research Center, Chinese Knowledge Center of Engineering Science and Technology (CKCEST), Engineering Research Center of Digital Library, Ministry of Education.
	
	%
	\bibliographystyle{ACM-Reference-Format}
	\balance
	\bibliography{sample-sigconf}
	
	%

\end{document}